# Loss-Aware Feature-Map Pruning in Convolutional Neural Networks Using Multi-Armed Bandits

Salem Ameen and Sunil Vadera

School of Science, Engineering and Environment, University of Salford, Salford, United Kingdom

*Corresponding author: Salem Ameen, s.a.ameen1@salford.ac.uk*

## Abstract

Convolutional neural networks often contain redundant feature maps that increase storage and inference cost without providing proportional gains in predictive performance. Feature-map pruning is a structured compression strategy because it removes complete convolutional output channels and their producing filters, rather than isolated scalar weights. The study extends the previously published MAB weight-pruning framework from unstructured scalar-weight removal to structured convolutional feature-map pruning, where removing one selected map also removes its producing filter and the corresponding next-layer input-channel kernels. The proposed method is a fixed-budget single-arm bandit search followed by top-k structured pruning. At each play time, one candidate feature map is temporarily masked and evaluated on a sampled mini-batch. The map is then restored, and the observed loss change is converted into a reward. After the play budget is exhausted, all candidate feature maps are ranked by their learned safe-removal scores, and the top-k maps are permanently removed together with their producing filters, biases and corresponding next-layer input-channel kernels. The paper evaluates UCB1 and Thompson Sampling because both policies allocate the evaluation budget adaptively across candidate feature maps and produce final scores suitable for top-k selection. The methods are first compared with direct/oracle-style feature-map evaluation on LeNet trained on MNIST, where the learned rewards show strong agreement with the direct method. The evaluation is then extended to MNIST, CIFAR-10, CIFAR-100, SVHN, CUB-200-2011 and Oxford Flowers 102. The results show that UCB1 and Thompson Sampling preserve accuracy close to the original unpruned models while removing feature maps and reducing convolutional computation. Friedman and Nemenyi tests show statistically significant differences between methods: UCB1 obtains the highest mean rank, followed by Thompson Sampling, and both proposed methods significantly outperform greedy and magnitude-based pruning while remaining statistically comparable to the original unpruned model.

## 1. Introduction

Convolutional neural networks (ConvNets) have become central to image classification and recognition tasks, but their predictive performance is often obtained by using many filters, feature maps and parameters [13], [14]. Such models can be expensive to store and evaluate, particularly when they are deployed on embedded, mobile or real-time systems [2], [10]-[12].

Pruning is a direct approach for reducing a trained network after learning [2]-[4]. In ConvNets, pruning can be applied at several levels, including individual weights, filters, channels and feature maps [10]-[12]. Removing individual weights may produce irregular sparsity, whereas feature-map pruning removes complete convolutional outputs and their producing filters. This gives a structured reduction that can reduce dense tensor dimensions and FLOPs [10]-[12].

The main difficulty is deciding which feature maps can be removed safely. Direct or oracle-style pruning estimates importance by removing each candidate feature map in turn and measuring the resulting change in loss, but this becomes expensive for large data sets and wide convolutional layers [2], [10]. Simple heuristics, such as activation statistics or filter magnitude, are cheaper but may not measure the functional effect of removal on the loss [10]-[12].

This paper addresses the trade-off between accurate importance estimation and computational cost by formulating feature-map pruning as a fixed-budget multi-armed bandit problem [1], [5], [6]. Each candidate feature map is an arm. At each play time, the bandit policy selects one feature map, temporarily masks it, evaluates the loss change, restores it and updates the selected arm statistics. The search phase is therefore a single-arm MAB process. The final pruning phase is separate: after the budget finishes, the method selects the top-k feature maps with the highest learned safe-removal scores and removes them structurally.

This article extends the previously published MAB weight-pruning framework [1] to a different structured-pruning problem: convolutional feature-map pruning. In the earlier work, each arm represented an individual scalar weight, and selected weights were set to zero. In the present method, each arm represents a complete feature map, and the final top-k selection is converted into structural deletion of filters, biases and corresponding next-layer input-channel kernels.

The main contributions are as follows:

1. A formal formulation of convolutional feature-map pruning as a fixed-budget top-k selection problem, where each MAB play evaluates one candidate feature map and the final pruning decision removes the top-k maps with the highest learned safe-removal rewards.
2. A loss-aware temporary masking procedure for estimating feature-map removability without permanently changing the network during the search phase.
3. UCB1 and Thompson Sampling feature-map pruning algorithms that estimate safe-removal scores from temporary masking trials and select the top-k feature maps for permanent structured pruning.
4. A structural deletion rule that removes producing filters, biases and corresponding next-layer input-channel kernels, including the case where the next layer is fully connected.
5. A FLOPs analysis that accounts for current-layer reduction, next-layer input-channel reduction and joint all-layer pruning.
6. An empirical evaluation on MNIST, CIFAR-10, CIFAR-100, SVHN, CUB-200-2011 and Oxford Flowers 102, including statistical comparison using Friedman and Nemenyi tests.

## 2. Related work

Pruning methods differ mainly in how they estimate the importance of the unit being removed [2]-[4]. Magnitude-based methods remove weights, filters or channels with small norms [10]. They are simple and computationally inexpensive, but small magnitude does not necessarily imply a small effect on the loss [4]. Second-order sensitivity methods such as Optimal Brain

Damage and Optimal Brain Surgeon use curvature information, but they are computationally demanding for modern networks [3], [4].

Activation-based feature-map pruning evaluates the behaviour of the produced activations [11], [12]. For example, a feature map with low activation variance may act similarly to a weak or redundant component, while a feature map with a high percentage of zero activations after ReLU may be less informative [11], [12]. These criteria are useful because they can be estimated with forward passes, but they remain indirect measures of the loss change caused by structural deletion.

Direct feature-map pruning measures importance by temporarily deleting each feature map and recording the corresponding change in loss [2]. It is loss-aware and provides a strong oracle-style reference, but it is expensive because it evaluates every candidate map, often over many examples.

Multi-armed bandits provide a principled framework for allocating a limited evaluation budget across uncertain alternatives [5], [6]. UCB1 uses optimism under uncertainty by combining empirical reward with an exploration bonus [5]. Thompson Sampling uses posterior sampling and can continue exploring arms whose success probabilities remain uncertain [6].

The method retains the loss-awareness of direct pruning while reducing exhaustive full-data candidate evaluation through adaptive sampling [1], [5], [6].

# 3. Proposed method

The method is formulated as follows. The key distinction is between the search phase and the final pruning phase. During the search phase, one feature map is temporarily evaluated at each play time. No feature map is permanently removed during this phase. After the fixed play budget is exhausted, the feature maps are ranked by their learned safe-removal scores, and the top-k maps are permanently removed.

## 3.1 Convolutional feature-map notation

Let $f_\theta$ be a trained ConvNet with parameters $\theta$. For a convolutional layer $l$, let $X^{(l-1)}$ denote the input tensor and $X^{(l)}$ denote the output activation tensor after nonlinearity. Let $C_{l-1}$ be the number of input channels and $C_l$ the number of output feature maps. The convolutional filters of layer $l$ are denoted by $W^{(l)}$, and the bias vector is denoted by $b^{(l)}$. For a square kernel of size $k_l \times k_l$, the filter tensor has shape

$$W^{(l)} \in \mathbb{R}^{C_l \times C_{l-1} \times k_l \times k_l}, b^{(l)} \in \mathbb{R}^{C_l}.$$

For output feature map $c$ at spatial position $(u, v)$, the pre-activation value can be written as

$$Z_{c,u,v}^{(l)} = b_c^{(l)} + \sum_{d=1}^{C_{l-1}} \sum_{r=1}^{k_l} \sum_{s=1}^{k_l} W_{c,d,r,s}^{(l)} X_{d,u+r,v+s}^{(l-1)}.$$

After applying the activation function $\sigma_l$, the output feature map is

$$X_c^{(l)} = \sigma_l\left(Z_c^{(l)}\right), c = 1, \ldots, C_l.$$

Let $\mathcal{L}$ denote the set of convolutional layers considered for pruning. The candidate arms are the feature maps in these layers:

$$A = \{(l, c) : l \in \mathcal{L}, c = 1, \ldots, C_l\}.$$

This arm set is used throughout the bandit search and in the final top-k selection rule.

The number of candidate arms is

$$K = |A| = \sum_{l \in \mathscr{L}} C_l .$$

If all convolutional layers are pruned jointly, $A$ contains all candidate feature maps across those layers. If a single layer is considered, $A$ contains only the feature maps of that layer.

## 3.2 Temporary feature-map masking

During the search phase, temporary feature-map removal is represented by a binary mask for the selected layer:

$$m^{(l)} \in \{0,1\}^{C_l} .$$

The masked activation tensor is defined componentwise by

$$\widetilde{X}_c^{(l)}(x; m^{(l)}) = m_c^{(l)} X_c^{(l)}(x), c = 1, \ldots, C_l .$$

The unpruned layer corresponds to the all-one mask

$$m^{(l)} = 1_{C_l} .$$

To test the temporary removal of feature map $c$ in layer $l$, the single-map mask is

$$m^{(l,-c)} = 1_{C_l} - e_c ,$$

where $e_c$ is the $c$-th standard basis vector. If $a = (l, c)$, the temporarily masked network is denoted by

$$f_\theta(x; m^{-a}) = f_\theta(x; m^{(l,-c)}) .$$

At play time $t$, the bandit policy selects one arm only:

$$a_t = (l_t, c_t) \in A .$$

The selected feature map is temporarily masked, the loss is evaluated, and the mask is then removed. This temporary operation is used only to estimate feature-map importance. It is not structural pruning. Permanent deletion is performed only after the fixed play budget has been exhausted.

## 3.3 Loss change and reward definition

At round $t$, a mini-batch $B_t$ is sampled. The mini-batch loss under a mask is

$$L_{B_t}(m) = \frac{1}{|B_t|} \sum_{(x_j, y_j) \in B_t} \ell(f_\theta(x_j; m), y_j),$$

where $\ell$ is the task loss, such as cross-entropy for image classification. For arm $a = (l, c)$, the loss change caused by temporarily removing feature map $c$ in layer $l$ is defined as

$$\Delta_a(t) = L_{B_t}(1_{C_l}) - L_{B_t}(m^{(l,-c)}) .$$

With this sign convention, $\Delta_a(t)>0$ means that temporary removal reduces the mini-batch loss, $\Delta_a(t)\approx 0$ means that the feature map has little measurable effect on the loss, and $\Delta_a(t)<0$ means that temporary removal increases the loss and is likely to harm performance.

For UCB1, a bounded reward is used:

$$r_a(t)=min\left\{1, max\left\{0, \frac{\tau+\Delta_a(t)}{s_r}\right\}\right\},$$

where tau $\tau$ is a tolerance parameter and $s_r$ is a scaling constant. The tolerance allows a small loss increase to be treated as acceptable, and the scaling constant keeps rewards in [0, 1], which is required by the bounded-reward form used with UCB-style selection [1], [5]. If no tolerated loss increase is intended, tau is set to zero.

For Thompson Sampling, a Bernoulli success/failure reward is used:

$$b_a(t)=1\{\Delta_a(t)+\tau\geq 0\}.$$

For Thompson Sampling, the Bernoulli reward uses tau = 0 unless a non-zero tolerance is explicitly reported as a hyperparameter. Changing tau changes the definition of a successful temporary removal.

Let $n_a(t)$ be the number of times arm $a$ has been selected up to round $t$ and let $\mu_a(t)$ be its empirical mean bounded reward. After selecting arm $a_t$ and observing reward $r_{a_t}(t)$, the count and empirical mean are updated as

$$n_{a_t}(t)=n_{a_t}(t-1)+1,$$

$$\mu_{a_t}(t)=\mu_{a_t}(t-1)+\frac{r_{a_t}(t)-\mu_{a_t}(t-1)}{n_{a_t}(t)}.$$

The empirical mean $\mu_a$ estimates how safely a feature map can be removed. A feature map with a high safe-removal score is a stronger candidate for final pruning.

## 3.4 Fixed-budget single-arm MAB framework for top-k feature-map pruning

Algorithm 1 gives the common framework. The policy-specific selection rule is supplied by UCB1 or Thompson Sampling [5], [6]. The algorithm evaluates one arm at each play time and performs top-k structural deletion only after all T plays have been completed.

Algorithm 1. Fixed-budget single-arm MAB framework for top-k feature-map pruning

**Input:** trained ConvNet $f_\theta$; data set $D$; candidate layer set $\mathcal{L}$; candidate arms $A$; play budget $T$; final pruning cardinality $k$; mini-batch sampler; reward parameters $\tau$ and $s_r$; bandit policy $\pi$.

**Output:** structurally pruned ConvNet $f_{\theta'}$.

1. Initialise $n_a=0$ and policy-specific score variables for every arm $a\in A$.
2. For $t=1,\ldots,T$, draw a mini-batch $B_t$ from $D$.
3. Select one arm $a_t=(l_t, c_t)$ using the current bandit policy.
4. Construct the temporary mask $m^{(l_t,-c_t)}=1_{C_{l_t}}-e_{c_t}$.

5. Compute the unmasked mini-batch loss $L_{B_t}\left(1_{C_{l_t}}\right)$.
6. Compute the temporarily masked mini-batch loss $L_{B_t}\left(m^{(l_t,-c_t)}\right)$.
7. Compute $\Delta_{a_t}(t)$ and convert it into the policy reward.
8. Remove the temporary mask; no structural pruning is performed during the search phase.
9. Update only the selected arm statistics and policy state.
10. After $T$ plays, compute the final safe-removal score $s_a$ for every arm.
11. Select $S_k = TopK_{a \in A}(s_a, k)$.
12. Permanently delete all feature maps in $S_k$, their producing filters, their biases and the corresponding next-layer input-channel kernels.
13. Return the structurally pruned network $f_{\theta'}$.

## 3.5 UCB1 feature-map pruning

UCB1 selects the arm with the largest upper-confidence score [5]. For every arm already played at least once, the UCB1 score is

$$U_a(t) = \mu_a(t) + \sqrt{\frac{2\log(t)}{n_a(t)}}.$$

At round $t$, the selected arm is

$$a_t = \arg\max_{a \in A} U_a(t).$$

Unplayed arms are selected before the confidence score is used, so that each candidate feature map has at least one initial observation. After the play budget finishes, UCB1 uses the empirical mean reward as the final safe-removal score:

$$s_a^{UCB1} = \mu_a(T).$$

The final pruning set is therefore

$$S_k^{UCB1} = TopK_{a \in A}(\mu_a(T), k).$$

Algorithm 2. UCB1 selection and final top-k rule

1. If there exists an unplayed arm $a$ with $n_a = 0$, select one such arm.
2. Otherwise compute $U_a(t) = \mu_a(t) + \sqrt{2\log(t)/n_a(t)}$ for every arm $a \in A$.
3. Select $a_t = \arg\max_{a \in A} U_a(t)$.
4. After reward observation, update $n_{a_t}$ and $\mu_{a_t}$.
5. After $T$ plays, select $S_k^{UCB1} = TopK_{a \in A}(\mu_a(T), k)$.

## 3.6 Thompson Sampling feature-map pruning

Thompson Sampling uses a Bernoulli success/failure model [6]. To avoid confusion with the network $f_{\theta'}$, failure counts are denoted by $n_a^-$ rather than $f_a$. For each arm a, let $n_a^{+¿¿}$ be the number of successful temporary removals and $n_a^-$ be the number of failures. With a uniform Beta prior, the posterior distribution for the success probability of arm a is

$$p_a \mid n_a^{+¿, n_a^- \sim Beta ¿¿}$$

At each round, Thompson Sampling draws one sample from each posterior distribution:

$$\theta_a(t) \sim Beta ¿$$

The selected arm is

$$a_t = \underset{a \in A}{arg\,max}\, \theta_a(t).$$

After observing binary reward $b_{a_t}(t)$, the posterior parameters are updated by

$$n_{a_t}^{+¿ \leftarrow n_{a_t}^{+¿+b_{a_t}(t),¿} ¿}$$

$$n_{a_t}^{-} \leftarrow n_{a_t}^{-} + 1 - b_{a_t}(t).$$

After the play budget finishes, the posterior mean is used as the final safe-removal score:

$$s_a^{TS} = \frac{n_a^{+¿+1}}{n_a^{+¿+n_a^{-}+2}.¿} ¿$$

The final pruning set is

$$S_k^{TS} = TopK_{a \in A}(s_a^{TS}, k).$$

Algorithm 3. Thompson Sampling selection and final top-k rule

1. For each arm $a \in A$, draw $\theta_a(t) \sim Beta ¿$.
2. Select $a_t = arg\,max_{a \in A} \theta_a(t)$.
3. Temporarily mask the selected feature map and observe the binary reward $b_{a_t}(t)$.
4. Update $n_{a_t}^{+¿¿}$ and $n_{a_t}^{-}$ using the observed success/failure outcome.
5. After $T$ plays, compute $s_a^{TS} = ¿$.
6. Select $S_k^{TS} = TopK_{a \in A}(s_a^{TS}, k)$.

## 3.7 Final structural deletion rule

Let $S_k$ denote the final set of feature-map arms selected after the play budget has been exhausted. The set is obtained from the final safe-removal scores:

$$S_k = TopK_{a \in A}(\rho_a, k).$$

Here, $\rho_a$ is the final safe-removal score for arm $a$, and $k$ is the total number of feature maps to be permanently removed. For UCB1, $\rho_a$ is the empirical mean reward at the end of the search. For Thompson Sampling, $\rho_a$ is the posterior mean success probability. During the search phase, only one feature map is temporarily masked at each play time. Permanent deletion is performed only after the final set $S_k$ has been formed.

For each convolutional layer $l$, define the subset of selected feature-map indices in that layer as

$$S_l = \{c : (l, c) \in S_k\}.$$

The number of selected feature maps from layer $l$ is

$$q_l = |S_l|.$$

To make the deletion rule precise, let the convolutional weight tensor and bias vector of layer $l$ be

$$W^{(l)} \in \mathbb{R}^{C_l \times C_{l-1} \times k_l \times k_l}, b^{(l)} \in \mathbb{R}^{C_l}.$$

In this convention, the first dimension of $W^{(l)}$ indexes output feature maps, the second dimension indexes input channels, and $k_l \times k_l$ is the spatial kernel size.

For every selected arm $a = (l, c) \in S_k$, the producing filter and its bias are deleted:

$$W^{(l)}_{c,:,:,:}, b^{(l)}_c.$$

Equivalently, when several selected maps belong to the same layer, all filters and biases indexed by $S_l$ are deleted together:

$$W^{(l)}_{S_l,:,:,:}, b^{(l)}_{S_l}.$$

If the following layer is convolutional, the corresponding input-channel kernels in layer $l+1$ are also deleted:

$$W^{(l+1)}_{:,S_l,:,:}.$$

This operation removes the next-layer kernels that receive input from the feature maps deleted in layer $l$. If the following layer is fully connected, the block of fully connected weights connected to the flattened spatial positions of the removed feature maps is deleted instead.

If $\theta$ denotes the original parameter set, the remaining parameters after deletion are denoted by $\theta'$, and the structurally pruned model is

$$f_{\theta'}.$$

This rule converts the temporary masking used during the MAB search into a physically smaller convolutional network. The final model is therefore structurally pruned rather than only masked.

Figure 1 illustrates the atomic structured deletion operation used when a selected feature map is permanently pruned.

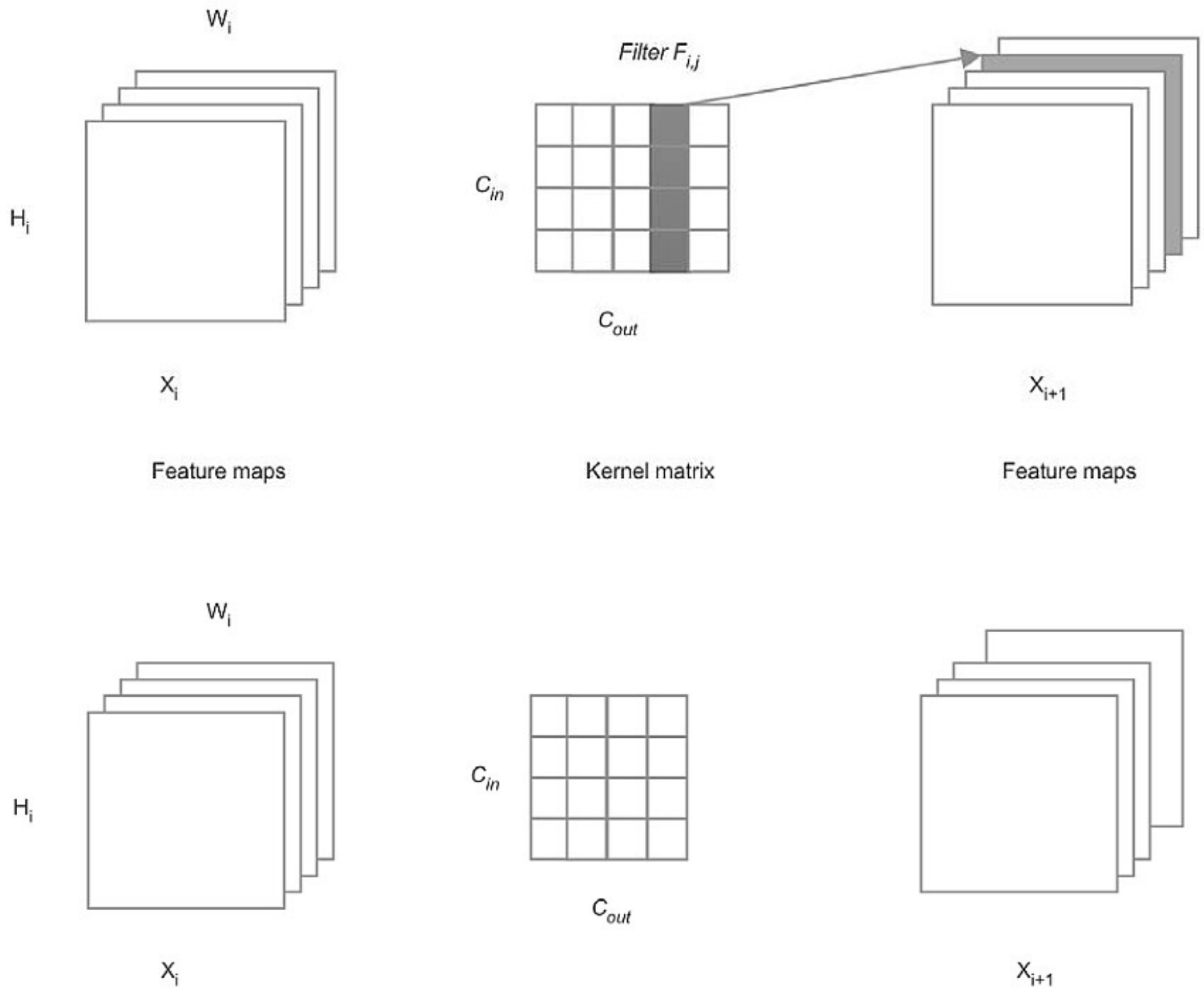

*Figure 1. Atomic structured deletion operation for one selected feature map. Removing filter F_{i,j} removes its corresponding output feature map X_{i+1,j}. When the final set S_k contains several selected maps, the same deletion operation is applied to every selected map.*

## 3.8 FLOPs reduction and computational complexity

For a convolutional layer $l$ with output spatial size $H_l \times W_l$, input-channel count $C_{l-1}$, output-channel count $C_l$ and kernel size $k_l \times k_l$, the approximate number of floating-point operations is

$$F_l = 2 H_l W_l \left( C_{l-1} k_l^2 + 1 \right) C_l .$$

The $+1$ term accounts for the bias operation. If $q_l$ feature maps are selected for deletion from layer $l$, then

$$C_l' = C_l - q_l .$$

The current-layer FLOPs after deleting $q_l$ output feature maps are

$$F_l' = 2 H_l W_l \left( C_{l-1} k_l^2 + 1 \right) \left( C_l - q_l \right).$$

The current-layer reduction is therefore

$$\Delta F_l = 2 H_l W_l \left( C_{l-1} k_l^2 + 1 \right) q_l .$$

If the next layer is convolutional, removing $q_l$ output maps from layer $l$ also removes $q_l$ input channels from layer $l+1$. The next-layer reduction caused by this input-channel deletion is

$$\Delta F_{l+1} = 2 H_{l+1} W_{l+1} q_l k_{l+1}^2 C_{l+1} .$$

No $+1$ bias term is included in this next-layer reduction because removing input channels deletes kernel multiplications, not output bias operations.

When pruning all convolutional layers jointly, a more precise layer-wise expression accounts for feature maps removed in both the previous and current layers:

$$F_l' = 2 H_l W_l \left( \left( C_{l-1} - q_{l-1} \right) k_l^2 + 1 \right) \left( C_l - q_l \right).$$

Thus, feature-map pruning reduces the computation of the layer that produces the deleted feature maps and, when the following layer is convolutional, the computation of the following layer as well.

Let $C_F$ denote the cost of one forward pass on a mini-batch, $T$ the play budget, $K$ the number of candidate arms, $B$ the mini-batch size and $C_\pi(K)$ the cost of selecting an arm under policy $\pi$. Each MAB play requires one forward pass for the unmasked model and one forward pass for the temporarily masked model. Therefore,

$$C_{MAB} = O\left( T \left( 2 C_F + C_\pi(K) \right) \right).$$

For straightforward UCB1 or Thompson Sampling scoring, the policy-selection cost is linear in $K$:

$$C_\pi(K) = O(K).$$

Thus,

$$C_{MAB} = O\left( T \left( 2 C_F + K \right) \right).$$

Direct evaluation over the full data set must evaluate all $K$ candidate feature maps. Its cost is approximately

$$C_{direct} = O\left( K \frac{N}{B} C_F \right).$$

The MAB search is more efficient than exhaustive full-data direct evaluation when

$$T \left( 2 C_F + K \right) \ll K \frac{N}{B} C_F.$$

When the forward-pass cost dominates policy-selection cost, this condition simplifies to

$$T \ll K \frac{N}{B}.$$

The factor $K$ must remain on the direct-evaluation side because direct pruning evaluates all candidate feature maps, not only one map.

### 3.9 Statistical comparison

The empirical comparisons are assessed using the Friedman test followed by the Nemenyi post-hoc test [7]-[9]. Let $K_m$ be the number of methods and $N_d$ be the number of data sets or experimental cases. Let $r_{ij}$ be the rank of method j on data set i. The average rank of method j is

$$R_j = \frac{1}{N_d} \sum_{i=1}^{N_d} r_{ij}.$$

The Friedman statistic is

$$\chi_F^2 = \frac{12 N_d}{K_m(K_m+1)} \left[ \sum_{j=1}^{K_m} R_j^2 - \frac{K_m(K_m+1)^2}{4} \right].$$

When the Friedman test rejects the null hypothesis, the Nemenyi test compares average ranks using the critical difference

$$CD = q_{\alpha, K_m} \sqrt{\frac{K_m(K_m+1)}{6 N_d}}.$$

Here $q_{\alpha, K_m}$ is obtained from the Studentized range distribution [7]-[9]. Differences in mean rank greater than CD are considered statistically significant at significance level alpha.

## 4. Experimental methodology

The evaluation has two parts. First, the proposed UCB1 and Thompson Sampling pruning methods are compared with the direct method on LeNet trained on MNIST [13]. This experiment tests whether the reward estimates learned by the MAB policies agree with direct loss-based feature-map evaluation. Second, the proposed methods are evaluated across several image-classification benchmarks and compared with the original unpruned model, greedy pruning and magnitude-based pruning.

The data sets are MNIST [13], CIFAR-10, CIFAR-100 [15], SVHN [16], Caltech-UCSD Birds 200-2011 [17] and Oxford Flowers 102 [18]. LeNet-style and ConvNet architectures are used for the smaller data sets [13]. For Birds-200 and Oxford Flowers 102, AlexNet is used in a transfer-learning setting after pre-training on ImageNet [14], [19].

For Birds-200, AlexNet is first pre-trained on ImageNet and used as a feature extractor [14], [17], [19]. The 1000-class output layer is replaced with a 200-class output layer. The new layer is trained using stochastic gradient descent with momentum 0.9, batch size 64, learning rate 0.001 and weight decay 0.0001 for 90 epochs while the earlier AlexNet weights are fixed. The whole network is then fine-tuned with learning rate 0.0001 for 10 epochs. For Oxford Flowers 102, the procedure is similar, except that AlexNet is fine-tuned for 40 epochs with learning rate 0.00001 [18].

The proposed methods are compared with the original unpruned model, greedy feature-map pruning and magnitude-based filter pruning [10]-[12]. In the main experiments, the play budget is set to five times the number of candidate feature maps. For example, if the candidate layer has 256 feature maps, the play budget is 1280 trials. When all convolutional layers are considered jointly, the arm set contains all feature maps across those layers, and the final top-k set may contain maps from different layers.

The direct method is used only as an oracle-style reference in the initial LeNet/MNIST comparison [2]. Direct pruning removes each candidate feature map in turn and records the loss change. The MAB methods are expected to approximate this direct ranking while evaluating only one selected feature map at each play time.

## 5. Results

### 5.1 Agreement with the direct method

The initial LeNet/MNIST comparison shows that UCB1 and Thompson Sampling learn reward patterns that are strongly aligned with the direct method. The Pearson correlation coefficient between the direct method and UCB1 is 0.83, and the correlation between the direct method and Thompson Sampling is 0.80. This indicates that the proposed MAB methods approximate the direct feature-map evaluation while avoiding exhaustive full-data assessment of every candidate feature map.

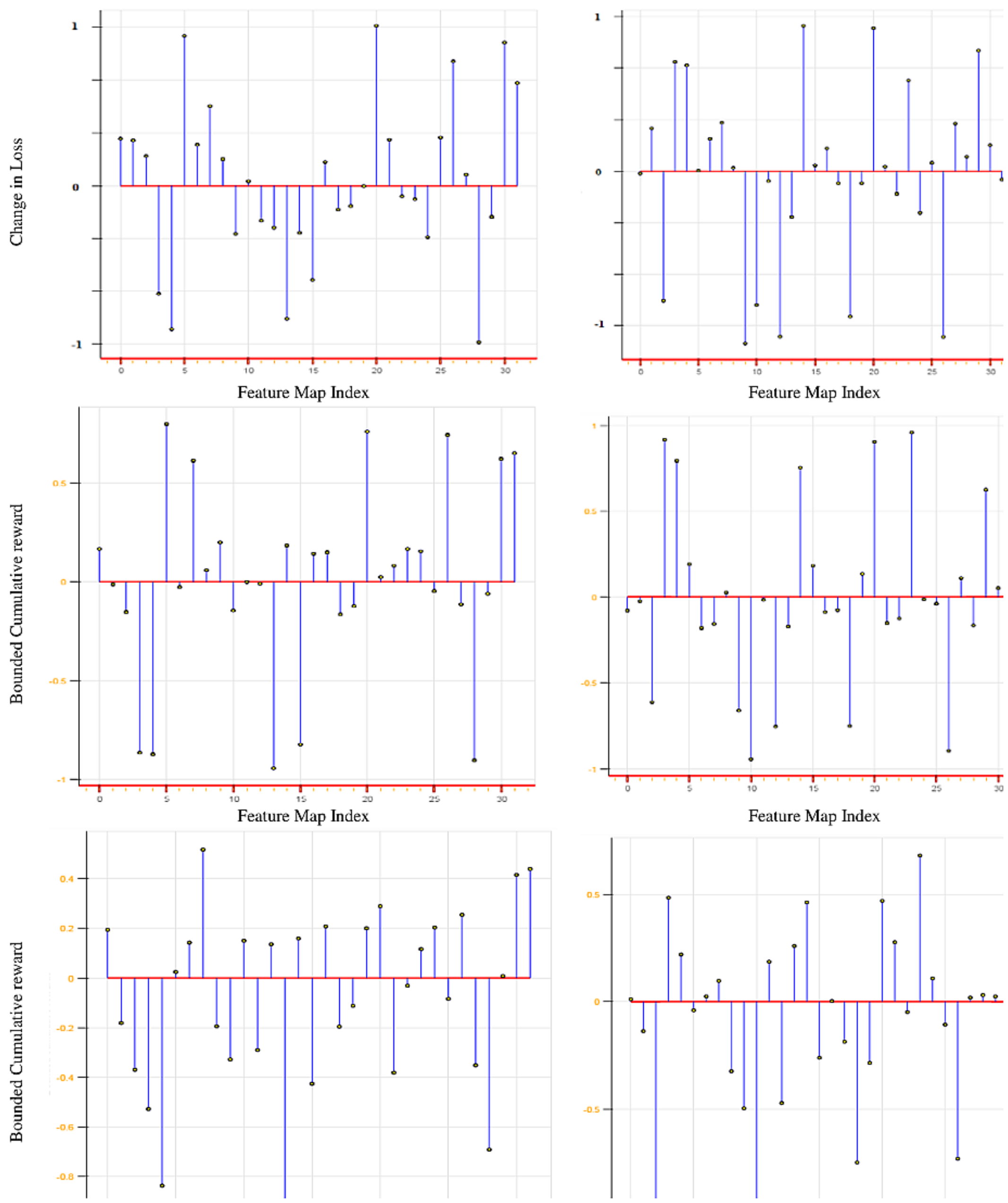

Figure 2. Direct evaluation, UCB1 and Thompson Sampling for temporary removal of individual feature maps in LeNet/MNIST. Each point corresponds to the effect or learned reward of removing one candidate feature map. These scores are used to rank maps, and the final pruning step removes the top-k candidates.

Figure 2 compares the loss-based direct evaluation with the safe-removal scores learned by UCB1 and Thompson Sampling for individual feature maps in LeNet/MNIST. Each point corresponds to one candidate feature map. The direct method evaluates the effect of temporarily removing that map, whereas the MAB methods estimate its removability through repeated sampled trials. The similar patterns across the plots show that the bandit policies assign high scores to many of the same feature maps identified by the direct method. This supports the use of UCB1 and Thompson Sampling as adaptive approximations to exhaustive direct feature-map evaluation.

The cumulative regret comparison also supports this interpretation. Compared with the greedy selection strategy, UCB1 and Thompson Sampling show lower regret relative to the direct method. Each point in the direct-method comparison corresponds to temporary evaluation of one candidate feature map. These scores are used to rank maps, and the final pruning step removes the top-k candidates after the play budget.

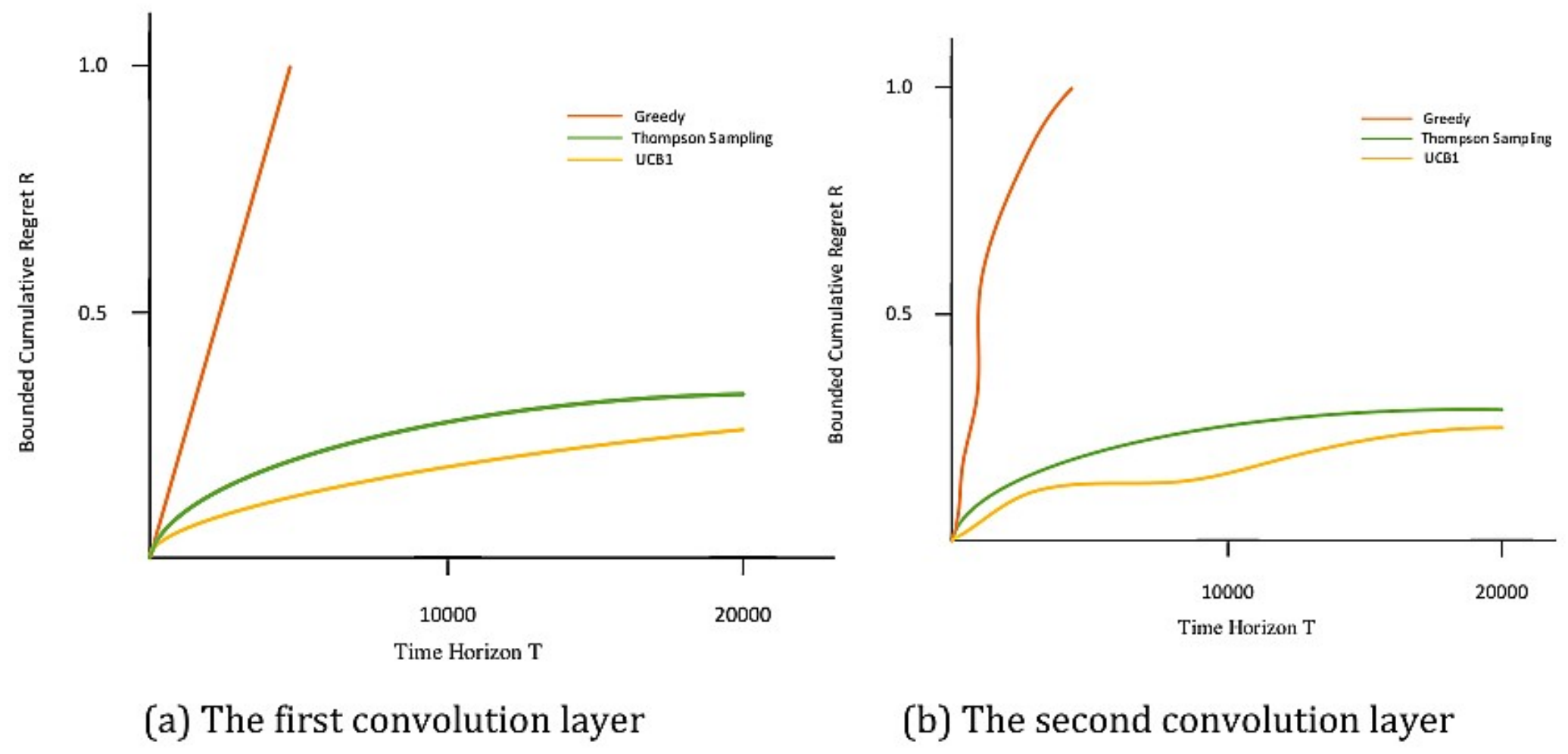

*Figure 3. Cumulative regret of UCB1 and Thompson Sampling compared with the direct method on the first and second convolutional layers of LeNet/MNIST.*

## 5.2 Accuracy after feature-map pruning

Table 1 gives a compact version of the feature-map pruning results. The table reports the original model accuracy, the accuracy after greedy pruning, magnitude-based pruning, UCB1 pruning and Thompson Sampling pruning, together with the remaining number of feature maps. These baselines follow common activation- or magnitude-based structured-pruning criteria used in earlier feature-map and channel-pruning work [10]-[12].

| Case | Original | Greedy | Mag. | UCB1 | TS | Maps left | Removed (%) |
|---|---|---|---|---|---|---|---|
| LN-MNIST L1 | 0.98 | 0.98 | 0.91 | 0.99 | 0.99 | 28 | 12.50 |
| LN-MNIST L2 | 0.98 | 0.97 | 0.90 | 0.98 | 0.98 | 26 | 18.75 |
| LN-MNIST All | 0.98 | 0.98 | 0.91 | 0.99 | 0.99 | 50 | 21.88 |
| C10 L1 | 0.81 | 0.80 | 0.73 | 0.81 | 0.81 | 28 | 12.50 |
| C10 L2 | 0.81 | 0.80 | 0.74 | 0.81 | 0.81 | 24 | 25.00 |
| C10 L3 | 0.81 | 0.80 | 0.73 | 0.81 | 0.81 | 52 | 18.75 |
| C10 L4 | 0.81 | 0.80 | 0.73 | 0.81 | 0.81 | 50 | 21.88 |
| C10 All | 0.81 | 0.80 | 0.74 | 0.82 | 0.82 | 140 | 27.08 |
| C100 L1 | 0.43 | 0.40 | 0.31 | 0.43 | 0.43 | 30 | 6.25 |
| C100 L2 | 0.43 | 0.41 | 0.30 | 0.43 | 0.43 | 28 | 12.50 |
| C100 L3 | 0.43 | 0.42 | 0.35 | 0.43 | 0.43 | 56 | 12.50 |
| C100 L4 | 0.43 | 0.41 | 0.35 | 0.43 | 0.43 | 54 | 15.63 |
| C100 All | 0.43 | 0.42 | 0.35 | 0.44 | 0.44 | 148 | 22.92 |
| SVHN L1 | 0.96 | 0.94 | 0.87 | 0.96 | 0.96 | 28 | 12.50 |
| SVHN L2 | 0.96 | 0.94 | 0.88 | 0.96 | 0.96 | 28 | 12.50 |
| SVHN L3 | 0.96 | 0.95 | 0.90 | 0.96 | 0.96 | 54 | 15.63 |
| SVHN L4 | 0.96 | 0.95 | 0.90 | 0.96 | 0.96 | 50 | 21.88 |

| Case | Original | Greedy | Mag. | UCB1 | TS | Maps left | Removed (%) |
|---|---|---|---|---|---|---|---|
| SVHN All | 0.96 | 0.95 | 0.91 | 0.97 | 0.96 | 140 | 27.08 |

Table 1a. Feature-map pruning results for LeNet and ConvNet experiments. LN = LeNet, C10 = CIFAR-10, C100 = CIFAR-100, Mag. = magnitude-based pruning, TS = Thompson Sampling.

| Case | Original | Greedy | Mag. | UCB1 | TS | Maps left | Removed (%) |
|---|---|---|---|---|---|---|---|
| F102 L1 | 0.80 | 0.76 | 0.74 | 0.80 | 0.80 | 90 | 6.25 |
| F102 L2 | 0.80 | 0.77 | 0.76 | 0.80 | 0.80 | 220 | 14.06 |
| F102 L3 | 0.80 | 0.78 | 0.76 | 0.81 | 0.82 | 340 | 11.46 |
| F102 L4 | 0.80 | 0.78 | 0.76 | 0.82 | 0.82 | 338 | 11.98 |
| F102 L5 | 0.80 | 0.78 | 0.77 | 0.83 | 0.82 | 228 | 10.94 |
| F102 All | 0.80 | 0.78 | 0.77 | 0.83 | 0.82 | 980 | 28.78 |
| B200 L1 | 0.71 | 0.68 | 0.64 | 0.71 | 0.71 | 90 | 6.25 |
| B200 L2 | 0.71 | 0.68 | 0.65 | 0.71 | 0.71 | 222 | 13.28 |
| B200 L3 | 0.71 | 0.69 | 0.67 | 0.72 | 0.71 | 344 | 10.42 |
| B200 L4 | 0.71 | 0.70 | 0.69 | 0.72 | 0.71 | 334 | 13.02 |
| B200 L5 | 0.71 | 0.70 | 0.69 | 0.72 | 0.72 | 226 | 11.72 |
| B200 All | 0.71 | 0.70 | 0.70 | 0.72 | 0.72 | 978 | 28.92 |

Table 1b. Feature-map pruning results for AlexNet transfer-learning experiments. F102 = Oxford Flowers 102, B200 = CUB-200-2011 Birds, Mag. = magnitude-based pruning, TS = Thompson Sampling.

The results show that UCB1 and Thompson Sampling generally maintain or improve the accuracy of the original models while pruning feature maps. The magnitude-based baseline is consistently weaker, and greedy pruning is less stable than the proposed MAB policies. This pattern is consistent with the loss-aware design of the proposed method and with the earlier evidence that MAB pruning can preserve network performance under a fixed evaluation budget [1].

## 5.3 Statistical comparison

The Friedman test indicates a statistically significant difference between the methods, with p = $4.25 \times 10^{-23}$. Table 2 reports the average ranks based on accuracy, where a higher rank is better [7]-[9].

| Method | Mean rank |
|---|---|
| UCB1 | 4.28 |
| Thompson Sampling | 4.10 |
| Original unpruned model | 3.55 |
| Greedy pruning | 2.03 |
| Magnitude-based pruning | 1.03 |

Table 2. Average ranks for feature-map pruning methods based on accuracy. Higher rank is better.

The Nemenyi post-hoc comparison gives a critical difference of CD = 1.133 at alpha = 0.05 [7]-[9]. UCB1 and Thompson Sampling are statistically better than greedy pruning and magnitude-based pruning because the differences in mean rank exceed the critical difference. The differences between the proposed methods and the original unpruned model do not exceed the critical difference, so the results do not provide evidence that the pruned models are statistically worse than the original model.

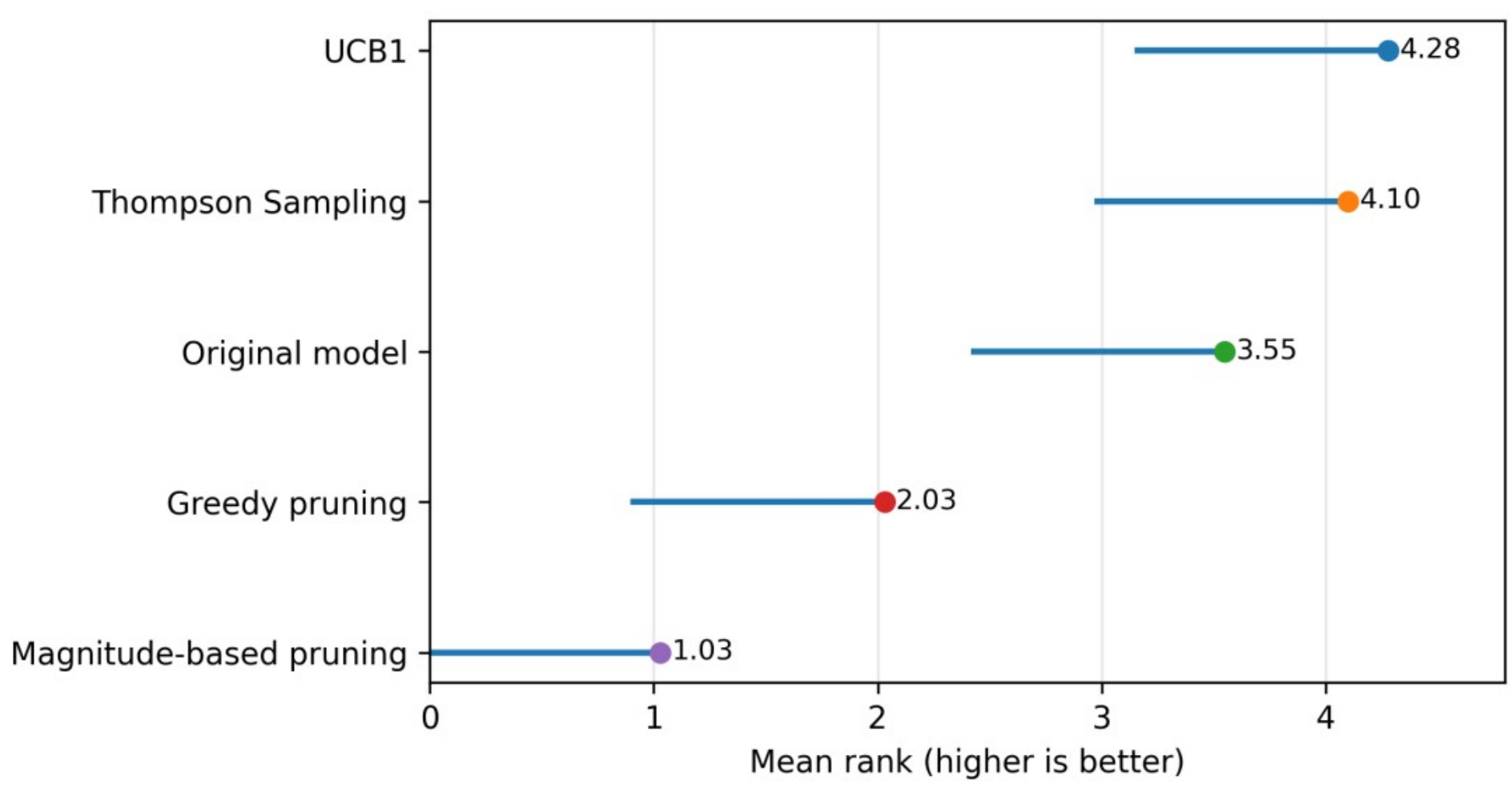

*Figure 4. Nemenyi post-hoc comparison for feature-map pruning. Dots show mean ranks, and horizontal segments have length equal to the critical difference CD = 1.133 at alpha = 0.05.*

### 5.4 Layer-wise and all-layer pruning

The experiments also compare pruning each convolutional layer separately with considering all convolutional layers jointly. Joint all-layer pruning is more flexible because the top-k set is not forced to remove the same proportion of feature maps from each layer. Instead, the bandit policy evaluates maps across the complete arm set and the final top-k set may contain feature maps from different layers. In the LeNet/MNIST all-layer setting, approximately 22% of the total feature maps are removed while maintaining or improving accuracy.

Later convolutional layers were expected to be more prunable because early layers tend to learn general features such as edges, while later layers capture more task-specific representations [14]. The experiments support this pattern in most ConvNet cases, but AlexNet behaves differently. A plausible explanation is that AlexNet was pre-trained on ImageNet, so some earlier-layer filters may be generic for ImageNet but less central for the target transfer-learning task [14], [19].

## 6. Discussion

The results indicate that UCB1 and Thompson Sampling are suitable policies for feature-map pruning because they allocate the evaluation budget adaptively [5], [6]. Direct pruning is loss-aware but computationally expensive because it evaluates every candidate map. Magnitude-based pruning is computationally inexpensive but does not directly measure the effect of feature-map removal on loss [10]. The proposed MAB methods provide a practical compromise because they learn removability from repeated temporary-removal trials.

A key methodological distinction is that the search and final pruning phases are distinct. During search, one feature map is evaluated at each play time and then restored. During final pruning, the top-k maps are deleted structurally. This distinction avoids confusing the method with multiple-play MAB, where several arms are selected within the same round. The method in this paper is single-arm MAB search followed by top-k structured pruning.

The statistical results support the usefulness of the method. UCB1 obtains the highest mean rank and Thompson Sampling obtains the second highest mean rank. Both proposed methods significantly outperform greedy and magnitude-based pruning while remaining statistically comparable to the original unpruned model. This outcome is desirable for compression because the model is reduced while accuracy is preserved.

Joint all-layer pruning is also important. When all convolutional layers are considered together, the final top-k set is not constrained to contain a fixed number of maps from each layer. This allows the method to identify redundancy wherever it occurs in the network and can avoid unnecessary pruning from sensitive layers.

## 7. Threats to validity

The evaluation covers several image classification data sets and architectures, but the results are conditioned on the specific trained models, pruning budgets and hyperparameters used. Different training schedules, fine-tuning procedures or pruning cardinalities may change the relative performance of the methods. The main metric is accuracy, while real deployment also depends on wall-clock latency, memory access patterns and hardware implementation. FLOPs reduction is an informative proxy for computational cost, but future work should measure actual inference time and energy consumption on target devices. The Thompson Sampling reward uses a binary success definition; if a tolerance parameter $\tau$ is introduced, it should be reported clearly because it affects which temporary removals are counted as successful. Finally, the experiments focus on post-training pruning; iterative prune-and-fine-tune schedules may yield additional gains.

## 8. Conclusion and future work

This study presented a loss-aware feature-map pruning framework based on multi-armed bandits. The proposed method is a fixed-budget single-arm MAB search followed by top-k structured pruning. Each play temporarily masks one candidate feature map, evaluates the resulting loss change, restores the map and updates the selected arm statistics. After the play budget finishes, the feature maps with the highest safe-removal scores are permanently removed together with their producing filters, biases and corresponding next-layer input-channel kernels.

The experiments show that UCB1 and Thompson Sampling approximate direct feature-map evaluation on LeNet/MNIST, preserve accuracy across several ConvNet benchmarks and outperform greedy and magnitude-based pruning in average rank. Statistical testing using Friedman and Nemenyi procedures confirms that the proposed methods significantly outperform the two pruning baselines while remaining statistically comparable to the original unpruned model [7]-[9].

Future work should evaluate the method on more recent architectures, including residual and transformer-based vision models; combine feature-map pruning with fine-tuning and iterative pruning schedules; and measure actual hardware latency, memory consumption and energy use in addition to FLOPs reduction.

## Declarations

**Conflict of interest:** The authors declare no conflict of interest.

**Funding: No external funding was received for this work.**

Data availability: The data sets used in this study are publicly available benchmark data sets: MNIST [13], CIFAR-10 and CIFAR-100 [15], SVHN [16], CUB-200-2011 [17] and Oxford Flowers 102 [18]. No private or restricted data were generated or analysed.

Code availability: The implementation materials associated with the pruning experiments are available at https://github.com/SalemAmeen/pruning_deep. The statistical-testing notebooks for the feature-map pruning comparisons are available at https://github.com/SalemAmeen/testing_python_friedman/tree/master/pruning_fearturemap/Inbound.

**Author contributions:** Salem Ameen developed the algorithms, implemented the experiments and drafted the manuscript. Sunil Vadera supervised the research, contributed to the design of the study and reviewed the manuscript.